\documentclass{article}
\usepackage[utf8]{inputenc}
\usepackage{mathpazo}
\usepackage{setspace}
\usepackage{indentfirst}
\usepackage{hyperref}
\usepackage{graphicx}
\usepackage[margin=1.25in]{geometry}
\usepackage{float}
\usepackage{caption}
\usepackage{subcaption}
\usepackage{lipsum}
\usepackage{vhistory} % for version history
\usepackage[numbers]{natbib}
\usepackage{amsmath}
\usepackage{multirow}

\begin{document}
%\nocite{*}
\begin{titlepage} 
	\newcommand{\HRule}{\rule{\linewidth}{0.5mm}}
	\center 
	
	\HRule
	\\[0.7cm]
	{\huge\bfseries Biological insights on grammar-structured mutations improve fitness and diversity}
	\\[0.25cm] 
	\HRule
	\\[2.5cm]
	
	\Large
    Stefano Tiso\\[0.1cm]
    \normalsize
    University of Groningen\\[0.3cm]

	\Large
    Pedro Carvalho\\[0.1cm]
    \normalsize
    University of Coimbra\\[0.3cm]

    \Large
    Nuno Lourenço\\[0.1cm]
    \normalsize
    University of Coimbra\\[0.3cm]
    
    \Large
    Penousal Machado\\[0.1cm]
    \normalsize
    University of Coimbra\\[0.3cm]
	\large

	%[0.7cm]
	%\Large
	%Solange Margarido\\[0.2cm]
	%\large
	%2014193105
	
	\vfill\vfill\vfill 
	{\large\today}
\end{titlepage}

%\doublespacing
\onehalfspacing

\section{Introduction}
%\subsection{Mutations in Grammatical Evolution}
Genetic Programming (GP) is an Evolutionary Algorithm (EA) that evolves programs to solve a given task.
A relevant branch of GP is Grammar-Guided Genetic Programming (GGGP) which uses a grammar to translate the representation (i.e., the genotype, a data structure representing the program) into an executable program (i.e., the phenotype or solution).
In GGGP, practitioners can include domain knowledge by designing the grammar to bias the evolutionary search toward certain solutions.
This advantage has led to successful applications of GGGP in many problem domains \cite{hemberg2008genr8,de2002automatic,gruau1995automatic}.

Despite these successes, GGGP can face difficulties exploring challenging solution spaces.
Some GGGP approaches are prone to poor locality \cite{rothlauf2006locality}, meaning that small changes in the genotype lead to drastic changes in the phenotype.
These problems are evident in tasks like the evolution of optimizers for neural network training.
Carvalho et al. \cite{Carvalho2022} use GGGP to evolve optimizers that compete with state-of-the-art human-designed solutions. 
While the result is favorable, the low average fitness of the population and the scarcity of reasonable solutions suggest that evolution is not exploring the solution space efficiently. 
Evolution appears to be randomly sampling the solution space (occasionally stumbling on promising genotypes) \cite{Whigham2015ExaminingEvolution}, rather than systematically improving upon incrementally fit solutions. 
Locality is often regarded as a matter of representation \cite{rothlauf2006locality}, but insights from evolutionary biology hint that variation operators may be part of the problem.

%GGGP often uses tree structures as an intermediate representation for solutions that share hierarchical and non-linear properties with natural systems \cite{Erwin2009TheNetworks,Yu2016TranslationSubsystems,Mengistu2015TheHierarchy}.
%Due to these properties, mutations can have vastly different effects on the phenotype depending on the gene they affect, with the ones that encode critical operations having a disproportionate effect on the phenotype.
In nature, different parts of the genome display different amounts of susceptibility to mutation \cite{Monroe2022MutationThaliana}.
Core genes that have a vital function tend to mutate less \cite{Martincorena2013Non-randomHypomutation}, while genes with minor effects tend to mutate more often \cite{Wray2007TheMutations}.
GGGP problems may share some of these dynamics with natural systems.
Nevertheless, GGGP traditionally applies variation operators (mutation and crossover) uniformly across all genes without differentiating their functional role \cite{mckay2010}.
By not discriminating among the different types of genes, researchers have little control over the effect of variation on the phenotype and, thus, on the evolutionary trajectory of the system.

In this work, we propose a method to overcome the uncontrolled effects of mutations and allow a more faithful inheritance of traits by mimicking biological systems.
We call our method Facilitated Mutation (FM), as it, in part, takes inspiration from the theory of Facilitated variation proposed by Kirschner and Gerhart in \cite{Gerhart2007}. 
In FM, each grammar non-terminal uses a different mutation rate according to its predicted influence on the phenotype: low-impact mutations are frequent, while high-impact mutations are rare.
We expect FM to enable the evolutionary algorithm to explore the solution space in a more structured way, reducing reliance on random sampling.
Note that FM is not a mutation operator, it is simply a novel way to apply existing mutation operators.
Consequently, the implementation cost of FM is minimal and the efficacy is related to the mutation operator used. 

We validate FM by evolving optimizers to train neural networks for image-recognition tasks. We compare our new mutational approach with the one used in past works \cite{Carvalho2022}.
Our contributions are as follows:
%\begin{enumerate}
%\item 
1) The best optimizers found by FM are statistically superior to the best optimizers found by competing approaches.
%\item 
2) The average fitness of the population is statistically superior using FM.
%\item 
3) FM statistically increases the diversity of discovered viable solutions.
%\item 
4) FM reduces the computational cost of experiments.% by directing search away from unviable solutions.
%\end{enumerate}

The remainder of this manuscript will be structured as follows:
In the Background, we provide the rationale from evolutionary biology guiding our design choices, as well as the current work in evolving optimizers and how it relates to larger GP and GGGP challenges.
In the Methods, we describe the implementation of FM and how it applies to optimizer evolution.
In the Experimental Setup, we describe our validation experiments and the metrics we use to compare FM against more traditional mutational approaches.
In the Results, we compare results from two versions of FM against two versions of traditional mutation approaches on four metrics: performance of the best solution, average performance of the evolved solutions, diversity, and computational cost.
%In the Results section, we will compare results from two versions of FM against two versions of a traditional mutation regime.
Finally, in the Conclusions section, we summarize our findings and extend our reasoning to general consideration and future directions.

\section{Background}
\subsection{Insights from Biological Evolutionary Theory}
Significant advancements in genomics, cell physiology, and developmental biology allowed evolutionary biology to gain an unprecedented understanding of the role of mutations and their effects on evolution. 
Of particular interest are insights on \textbf{heterogeneous mutation rates} and \textbf{heterogeneous mutation effects}.

%\subsubsection{Heterogeneous mutation rates}
Mutations that lead to phenotypic change do not happen homogeneously across the genome \cite{Feuk2006StructuralGenome,Gruber2012ContrastingDominance}.
There are conserved genes that code for fundamental \textit{core processes}, that only mutate their function  rarely, being more \textit{robust} to mutation.\cite{Campillos2006IdentificationNetworks,Tanay2005ConservationYeast}.
Many cellular and developmental processes are fundamentally unchanged since their emergence, such as the formation of micro-tubule structures \cite{Winther2001VarietiesBehaviors,Gerhart1997}, cell adhesion processes \cite{Tepass2001EpithelialDrosophila,Hartsock2008AdherensCytoskeleton,HutterConservationGenes}, and anteroposterior axis formation \cite{Gerhart2005HemichordatesChordates}.
Most mutations, instead, affect the phenotype by changing \textit{regulatory elements} \cite{Stern2009IsEvolution}.
Changes in these elements do not alter the function of the core processes but regulate their activities and how they combine \cite{Wagner2007TheModularity}.

%\subsubsection{Heterogeneous effects of mutations}
Not only do mutations not happen randomly across the genome, but they also have non-random effects on the phenotype depending on where they happen.
If a mutation alters the functioning of a core process, it will fundamentally change the inner workings of the organism, most likely leading to a non-viable phenotype.
Instead, mutations on regulatory sequences change the levels of expression of core processes and their interactions, producing functional phenotypes that are variations on the original theme.

Like biological systems, in evolutionary computation (EC) different genes play different roles in determining the final phenotype.
%more or less important 
%, being either more defining (core) or more fine-tuning (regulatory).
However, unlike biological systems, EC often 
%does not take this into account
%when implementing mutation
%, as it
implements only a single mutation rate for all genes.
By taking heterogeneous mutation rates and heterogeneous mutation effects into account EC systems might be able to display other interesting dynamics discussed by evolutionary biology: \textbf{developmental biases} and \textbf{facilitated variation}.

%\subsubsection{Developmental biases}
Empirical and theoretical research in evolutionary-developmental biology suggests that evolution has led to (and might have selected for) organisms whose core processes bias random mutations to produce a non-random subset of phenotypes.
In the literature, this phenomenon is called developmental bias \cite{Uller2018DevelopmentalPerspective,Watson2016}. 
Examples of developmental biases are: the number and distribution of digits, limbs, and segments in tetrapods \cite{Alberch1985AAMPHIBIANS}, the structural and pigment coloration of insect wing \cite{Brakefield2006ExploringBiology}, and flower morphology \cite{Wessinger2016AccessibilityEvolution}.

In line with the observations illustrated so far, the Theory of Facilitated Variation by Gerhart and Kirschner \cite{Gerhart2007} postulates that organisms are structured to produce variation easily and that this variation is biased towards adaptive phenotypes.
Organisms can do this because they contain an archive of evolved core processes that rarely change due to mutations and maintain functionality independently of context.
Core processes can easily combine their activities through mutations on regulatory sequences and, since they maintain functionality across contexts, the outcome of these combinations is biased to be neutral or beneficial. 

The predictions of this theory have profound implications for evolutionary dynamics.
Facilitated variation can explain how evolution is dramatically faster than we expect 
%without reducing its precision
since it selects core processes that favor the rapid discovery of functional solutions, instead of proceeding randomly \cite{Parter2008FacilitatedEnvironments}.
By applying heterogeneous mutation rates and considering the heterogeneous effects of mutations, EC systems
%can replicate some properties of natural evolution,
could behave in similar ways to what Facilitated variation predicts, possibly improving their efficiency and effectiveness.
By slowing the mutation rate of core genes and increasing the rate of regulatory ones, evolution would
%, EC system could behave in way similar to what Facilitated variation predicts.
%They would
explore only a few core processes at the time (low mutation rate), but in different combinations (high mutation rate). 
This would steer the search trajectory towards a reduced subset of all the possible solutions, giving time to select and improve the most successful ones before new ones are introduced.
Thus, EC systems could incrementally build an archive of good solutions that can recombine together effectively, granting a better exploration of the solution space.
%The slow mutation rate of core processes would then introduce gradually new potential core processes allowing anyway for further exploration of new combinations.

\subsection{Mutation and Crossover in GGGP}
In GGGP the genome of individuals encodes for a set of derived rules, defined by how its genome is translated by a grammar. 
These production rules can either be non-terminals (invoking more production rules) or terminals (constants or operations).
Mutations in GGGP change the genome, potentially changing one or more of the chained production rules. 

A change in a production rule can lead to an important reshaping of the resulting behavior. 
Mutating a terminal into a non-terminal (which will invoke more production rules) will expand the number of executed operations. 
%Each new operation is determined itself at the moment of mutation, potentially leading to new behavior. 
Mutating a non-terminal into a terminal instead will cancel a series of operations, preventing them from ever being called.
Thus, it is important to realize that, independently of which representation is used (e.g. Grammatical Evolution [GE], Structured Grammatical Evolution [SGE], etc.), GGGP mutations can cause different effects of different magnitudes depending on where they happen.
GGGP often uses mutations in tandem with another variation operator: crossover.
This is done to improve the exploration of the solution space.
During crossover, a region of the genotype of one individual is swapped with a region from another one. 
The resulting individual is therefore a novel combination of the two parent genotypes.
However, there is no guarantee that this new combination will resemble a combination of the two parents.
Traditionally in GGGP, crossover methods are operated not by identifying functionally equivalent regions of the genome, but by simply exchanging non-terminals.
Therefore, crossover in GGGP leads to highly diverse, but very unstructured variation.

As biology suggests, if core components are not protected, arbitrarily changing the genome structure (both through mutations and crossover) will likely lead to system-wide deleterious effects.
Nonetheless, studies in GGGP normally use a uniform rate of mutation over the entire genome and rely heavily on crossover.
%and set it to be almost intrinsic to reproduction (probability of crossover $\approx$ 0.9).

\subsection{Evolution of optimizers showcases both potential and issues of GGGP}

Optimizers are a relevant research topic in computer science as their application domains like Machine Learning, have deep practical implications.

These algorithms can train Artificial Neural Networks to solve a target task using back-propagation \cite{rumelhart1985learning}.
Recent research shows that GP is a viable way to design competitive optimizers.
GP systems such as PushGP\cite{lones2019,lones2020,lones2021} and AutoML-Zero\cite{chen2022} both have notable results in evolving optimizers.

GGGP also achieves important results through the AutoLR framework \cite{Carvalho2022}, which is based on the sub-branch of GGGP known as Structured Grammatical Evolution (SGE) \cite{Lourenco2016}.
AutoLR is  capable of evolving human-competitive optimizers from scratch without prior knowledge of the components used by human-made ones.
%This result is possible because of AutoLR's grammar, which guides the evolutionary search toward programs with the structure and operations of an optimizer.
However, despite the good results, some problems are evident in the way that AutoLR navigates the solution space.
Very few solutions obtain a high fitness and most discovered solutions are non-functional.
Moreover, high fitness solutions emerge mainly during early evolution and do not improve any further, suggesting that the evolutionary algorithm may be behaving similarly to random search\cite{Whigham2015ExaminingEvolution,Gottlieb2000AProblem,Gottlieb2000LocalityProblem}.
Random search can be helpful to explore a solution space \cite{Salimans2017,Gajewski2019EvolvabilityEvolvability,Bergstra2012}, but it does not fit the goals of Evolutionary Computation to incrementally produce diverse and competitive solutions.

The shortcomings of AutoLR are likely caused by the fragility of the solutions to homogeneous mutation rates and indiscriminate use of crossover.
Much like natural systems have core processes necessary for viable phenotypes, optimizers have core components (e.g., the gradient) that are likely to generate non-viable phenotypes when changed. 
But unlike natural systems, current optimizer evolution in GGGP does not consider heterogeneous mutation rates and mutation effects.
These issues are not exclusive to AutoLR and GGGP.
In fact, this is a well-known problem in GP \cite{Whigham2015ExaminingEvolution,Gottlieb2000AProblem,Gottlieb2000LocalityProblem}, where algorithms often display low locality (changes in the genotype are disproportionally reflected on the phenotype) \cite{Lourenco2016,Medvet2017}.

Some researchers have already worked to mitigate this issue.
Whigham \cite{Whigham1996} already in 1996 proposed the fine-tuning of heterogeneous mutation rates in GGGP. 
However, this is a small part of broader work from Whigham at the time, and only covers this idea briefly, not considering how biological evolution insights can inform design decisions in this subject.
There is also work that explores forms of heterogeneous mutation outside of GP.
Evolution Strategies (ES) are an entire type of EA that focuses heavily on adaptive control of parameters \cite{hansen2015evolution}.
These systems adapt mutation rates and mutation steps for more successful mutation outcomes.
However, these systems are commonly used for continuous search problems and do not address the same tasks as GGGP.
Additionally, ES do not have grammars that can easily link different genes to different mutation rates in a structured, expertly informed way. 
As such, the possible benefits of a solution that mimics the patterns of natural mutations to address poor locality remains largely unexplored.

\section{Methods}
\subsection{Facilitated Mutation}\label{sec:methods_facilitated mutation}
The traditional GGGP approach applies mutations% and crossover
uniformly across the genome %and does not exploit %the advantages of 
not exploiting more fine-tuned mutations as natural systems do. 
In this work, we propose a new mutation approach for GGGP, Facilitated Mutation (FM), to replicate %how
natural mutations% works
. 

Like in nature, mutations on different genes in GP produce different effects. FM uses heterogeneous mutation rates to support these differences.

%We apply these biological concepts to bring artificial evolution closer to producing developmental biases and Facilitated Variation as observed in nature.
The goal of this endeavor is to lead to an efficient, incremental search of the solution space, bringing tangible benefits.% to practitioners.

\begin{figure}[h]
\centering
\begin{subfigure}{0.3\columnwidth}
\centering
        \includegraphics[height=\textwidth]{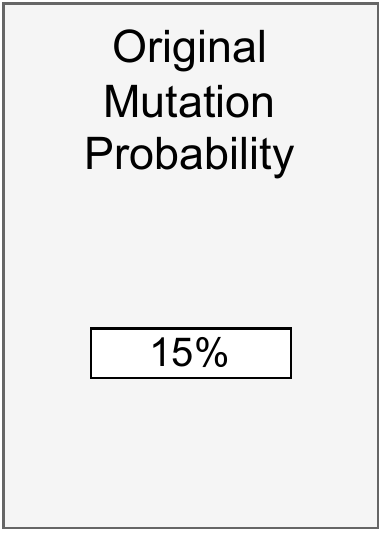}
        \caption{}
        \label{fig:a}
\end{subfigure}
\begin{subfigure}{0.3\columnwidth}
    \centering
    \includegraphics[height=\textwidth]{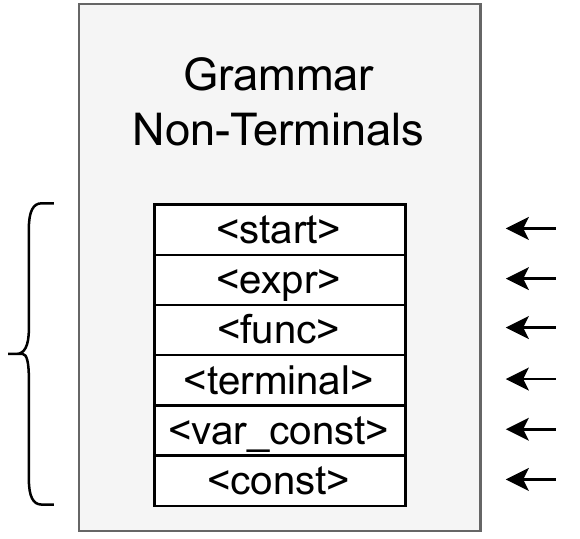}
        \caption{}
    \label{fig:b}
\end{subfigure}
\begin{subfigure}{0.3\columnwidth}
    \centering
    \includegraphics[height=\textwidth]{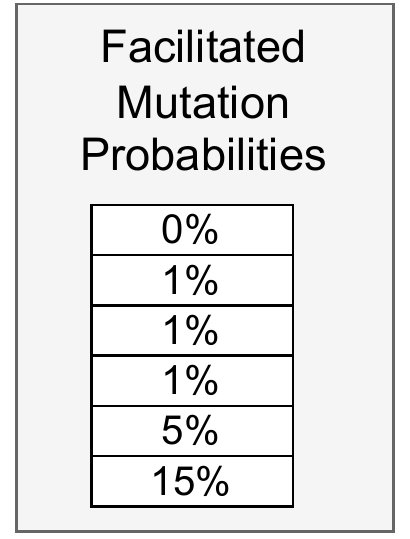}
    \centering
    \caption{}
    \label{fig:autolr_mutation_parameters}
\end{subfigure}
\caption{Traditionally, mutation operators use a single mutation rate (\ref{fig:a}) for all grammar non-terminals (\ref{fig:b}). In FM, there is a separate mutation rate for each non-terminal (\ref{fig:autolr_mutation_parameters})}

\end{figure}

Specifically, FM leverages the structure of the existing grammar to create several tiers of mutations.
Each grammar non-terminal uses a different mutation rate:
lower 
for more impactful non-terminals% rarely mutate
, %while 
higher for less influential ones% mutate more often
, as shown in Figure~\ref{fig:autolr_mutation_parameters}.
Separate mutation rates allow practitioners to use their expert knowledge to tune the rate of different mutation effects optimally.

It is important to notice that FM is not a mutation operator, the approach only regulates the application of the ones already in use.% for any specific problem.
FM is consequently widely applicable since it does not require the implementation of new complex algorithms or operators.
Instead, it only requires small adjustments to the way that mutation is applied in the system.

Moreover, biological insights suggest that swapping arbitrary regions of the genotype via crossover will more often than not disrupt the coordinated activity of genes, causing high-impact deleterious variation.
Thus FM, when it employs it, sets crossover at low rates to allow time for new combinations of genes to fine-tune their concerted action.

\subsection{Facilitated Mutation in Optimizer Evolution}
In this work, we apply the ideas of FM to AutoLR \cite{Carvalho2022}, a framework for the evolution of neural network optimizers based on a GGGP engine called SGE \cite{Lourenco2016}.
The evolution of optimizers is an excellent task to evaluate FM. 
Optimizers have clearly distinguishable core components (e.g., gradient) crucial to their functioning, and regulatory components (e.g., tuning parameter values) dedicate to their fine-tuning.
These distinct components would benefit from different mutation rates.

However, the original AutoLR grammar combines some crucial and tuning components in the same non-terminals.
Specifically, the gradient shares a non-terminal with other, less important ones. 
The gradient is the most important component of an optimizer, any change to the gradient in an active gene will likely result in a complete overhaul of the solutions' behavior.
To improve the efficacy of FM, we create a new grammar that separates components with different roles in different non-terminals.
%It is important to notice, however, that t
The new grammar
 still
 maintains the original grammar's functionality, as it uses the same components.
In Figure \ref{fig:old_new_grammar}, we show %an abridged version of 
the two grammars, side-by-side%, showcasing the changes made
.
The complete grammars can be found at \cite{autolr_repo_original_grammar,autolr_repo_facilitated_mutation_grammar}% are too large to include as most non-terminals are repeated for each custom variable and include a large number of productions.
We refer the reader to the AutoLR paper \cite{Carvalho2022} for details on the design decisions, operations, and variables of the original grammar.%, as well as specific about the variables and operations used.

Leveraging these new non-terminals, we set specific mutation rates based on expert knowledge.
We assign a 0.15 mutation rate to constants, allowing easy tuning of parameters.
We assign a 0.05 mutation rate for changes between constants and other variables.
We assign a 0.01 mutation rate to all other non-terminals.
This 0.01 mutation rate affects changes that influence the behavior of the optimizer the most i.e.: changes from the gradient to anything else, as well as changes in the operations between variables.   
Figure \ref{fig:autolr_mutation_parameters} presents the complete set of parameters.
Finally, we allow FM to make use of crossover (FM with crossover: FMX), but at a rate that is significantly smaller than previous setups (FMX crossover rate = 0.01 $<<$ original crossover rate = 0.9).
We chose this crossover rate to be just high enough to avoid stagnation on local minima but not high enough to disrupt solutions that emerge from mutation.

\begin{figure}[t]
\begin{subfigure}{0.45\columnwidth}
    \begin{align*}
        {<}\text{start}{>}::= & {<}\text{expr}{>}\\
        {<}\text{expr}{>}::= & {<}\text{func}{>}\,|\, {<}\text{term}{>}\\ 
        {<}\text{func}{>}::= & {<}\text{expr}{>}\text{+}{<}\text{expr}{>}\, |\\
        & \,{<}\text{expr}{>}\text{*}{<}\text{expr}{>} \, | \,\\& ...\\ 
        {<}\text{\underline{term}}{>}::= & {<}\underline{const}{>}\, |\\
        &\, \text{\underline{alpha}} \, | \, \text{grad} \, |\,\\& ...\\
        {<}\text{const}{>}::= & \, 0.0 \, | \, 5E^{-5} \, | \,\\& ...\\ 
    \end{align*}
    \caption{Original Grammar.}
    \label{fig:standard_example}
\end{subfigure}
\begin{subfigure}{0.45\columnwidth}
    \begin{align*}
        {<}\text{start}{>}::= & {<}\text{expr}{>}\\
        {<}\text{expr}{>}::= & {<}\text{func}{>}\,|\, {<}\text{term}{>}\\ 
        {<}\text{func}{>}::= & {<}\text{expr}{>}\text{+}{<}\text{expr}{>}\, |\\
        & \,{<}\text{expr}{>}\text{*}{<}\text{expr}{>} \, | \,\\& ...\\ 
        {<}\text{\underline{term}}{>}::= & {<}\text{\underline{var\_const}}{>}\, |\\
        &\, \text{grad} \, | \, ...\\
        {<}\text{\underline{var\_const}}{>}::= & {<}\text{\underline{const}}{>}\, |\\
        &\, \text{\underline{alpha}} \, | \,...\\
        {<}\text{const}{>}::= & \, 0.0 \, | \, 5E^{-5} \, | \, \\&...\\ 
    \end{align*}
    \caption{Grammar for Facilitated Mutation.}
    \label{fig:extended_example}
\end{subfigure}
\caption{The original grammar used in AutoLR (\ref{fig:standard_example}) and the adapted version for Facilitated Mutation (\ref{fig:extended_example}).
Non-terminals that have been added or changed are \underline{underlined}.}
\label{fig:old_new_grammar}

\end{figure}

\section{Experimental Setup} \label{sec:exp_setup}
%\subsection{Evolutionary simulation and mutational setups}
We validate Facilitated Mutation by using AutoLR to evolve optimizers that train a Convolutional Neural Network for image classification.
Optimizers are assigned a fitness value based on the classification accuracy achieved by the network they train%.Once fitness is assigned, optimizers
 and undergo tournament selection of size two.%: this algorithm compares two optimizers, and the one with the best fitness wins the tournament.
Winners of the tournament reproduce with a probability of undergoing mutation and crossover.% and then contribute to the next generation.
This procedure is repeated until the population for the next generation is filled up.
The population size is 100 individuals, and evolution runs for 200 generations.
The best solution of each generation is preserved via elitism %(copied directly into the next generation before tournament selection)
.
We run 30 independent simulations for four different mutation approaches: 
\begin{enumerate}
    \item\textbf{Facilitated Mutation with rare crossover (FMX)}: our novel solution implementing both changes to mutation rates and  to crossover (\textit{low rate}).
    \item\textbf{Facilitated Mutation (FM)}: our novel solution in isolation to investigate its effectiveness in the absence of crossover.
    \item\textbf{Original mutation without crossover (OM)}: the original solution without crossover, for direct comparisons with FM. 
    \item\textbf{Original mutation with crossover (OMX)}: the original setup used in past works \cite{Carvalho2022}.
\end{enumerate}
Table \ref{tab:evolution-parameters} describes the evolution parameters for each approach.
\begin{table}[t]
\centering
\caption{Evolutionary parameters for the four different setups}
\label{tab:evolution-parameters}
\begin{tabular}{|c|cccc|}
\hline
\multirow{2}{*}{\textbf{Parameters}} & 
\multicolumn{4}{c|}{Value} \\
\cline{2-5} & 
\multicolumn{1}{c|}{FMX}  &
\multicolumn{1}{c|}{FM} &
\multicolumn{1}{c|}{OM} &
OMX \\
\hline

Crossover Rate &
\multicolumn{1}{c|}{0.01} &
\multicolumn{1}{c|}{0.0} &
\multicolumn{1}{c|}{0.0} & 
0.9 \\
\hline
Mutation Rate & 
\multicolumn{2}{c|}{Heterogeneous} &
\multicolumn{2}{c|}{Homogeneous (0.15)} \\
\hline

Population Size & \multicolumn{4}{c|}{100} \\ \hline
Generations     & \multicolumn{4}{c|}{200} \\ \hline
Elitism         & \multicolumn{4}{c|}{1\%} \\ \hline
Tournament Size & \multicolumn{4}{c|}{2}   \\ \hline
Max Depth       & \multicolumn{4}{c|}{17}  \\ \hline
\end{tabular}\\
\end{table}
We use four metrics to analyze and compare the results from the different mutational approaches (FMX, FM, OM, OMX):
\begin{itemize}
    \item \textbf{Best Fitness}: The \textit{fitness of the best evolved solution} in a simulation.
    \item \textbf{Population Fitness}: The \textit{average fitness of the population}. We also consider the \textit{distribution of fitness in the population}.
    \item \textbf{Population Diversity}: The \textit{number of different viable and distinct solutions in a population}. We distinguish solutions by the equation %that describes the behavior of the optimizer they encode
    encoding the optimizer. We set a %n arbitrary
    threshold of accuracy% in the image classification executed by a network trained by a specific solution 
    to distinguish non-viable ($<$ 0.5) and viable ($>$ 0.5) solutions.
    %, and high-quality ($>$ 0.8) solutions.
    \item \textbf{Computational Cost}: \textit{Number of training events over the evolutionary run}. Most computational costs in AutoLR come from the %expensive network
    training necessary for the fitness evaluation.% of the evolving optimizers. 
    %The implementation of 
    AutoLR avoids re-evaluating solutions it has already encountered (see %subsection
    ~\ref{Solution archive and viable solution pre-selection}); thus, algorithms that bias variation can reduce computational costs.
\end{itemize}
\subsection{Task, training and fitness evaluation}
We evolve the optimizers to train a simple convolutional neural network \cite{autolr_repo} for image classification on Fashion-MNIST \cite{xiao2017fashion}.
Fashion-MNIST is a data set with 60000 28x28 gray-scale training images.
Each image depicts an article of clothing belonging to one of ten categories (e.g., T-shirt/top, trousers, pullover).
The optimizer trains the network to label these images correctly.
Training has two stop conditions. 
First, the procedure terminates when a maximum number of epochs is reached.
Second, an early stop mechanism terminates the procedure if there is no progress for a set number of consecutive epochs.
To ensure a fair fitness assessment, we split Fashion-MNIST into three data sets:
 Training Data - %Used to train the network.
    During training, the network assigns labels to this data and the optimizer changes the network's weights based on the errors of the assigned label. % Note that t
    The optimizer directly interacts with this data.
Validation Data - %Used to monitor training. 
    During training, validation metrics are calculated on this data to monitor progress. The early stop mechanism monitors these validation metrics to terminate training when no progress is made.
    The optimizer does not interact directly with this data.
Fitness Data - %Used to calculate fitness. 
    After training concludes, accuracy is calculated using this data and used as the optimizer's fitness. The optimizer does not directly interact with this data, ensuring an unbiased fitness assessment.

A table summarizing the parameter values used in our Fashion-MNIST experiments is presented in the supplementary material (Supplementary Material Table 1).
%\begin{table}[h]
%\centering
%\caption{Training parameters for optimizer fitness evaluation.}
%\label{tab:trainingparameters}
%\begin{tabular}{|c|c|}
%\hline
%\textbf{Parameters} & \textbf{Value}  \\ \hline
%Training Size       & 6500             \\ \hline
%Validation Size     & 3500             \\ \hline
%Fitness Size        & 50000             \\ \hline
%Training Epochs      & 100               \\ \hline
%Early Stop Metric   & Validation Loss \\ \hline
%Early Stop Patience & 5               \\ \hline
%\end{tabular}
%\end{table}

\subsection{Solution archive and viable solution pre-selection}
\label{Solution archive and viable solution pre-selection}
Once fitness is calculated, the optimizer and its fitness are recorded in an archive.
Before evaluating any optimizer, the system will check the archive for an existing record to avoid %redundant
re-evaluation, since multiple genomes can encode for the same optimizer.
Additionally, the system can detect dysfunctional optimizers before evaluation: %by searching for the gradient in the active genes. A
 an optimizer that does not use the gradient cannot train a neural network, so AutoLR assigns them a poor fitness value (0.1) and skips evaluation% to save computational resources.
These two procedures together save considerable computational resources. 
\subsection{Post-hoc Analysis}
While the fitness function provides a quick assessment of the quality of an optimizer, it is ultimately an abridged procedure that focuses on efficiency.
We use limited training size, few training epochs, and an early stop mechanism to speed up computation during the evolutionary simulations.
However, these expedients limit the thoroughness of the training and thus hinder the assessment of the full capabilities of an optimizer.
We perform a post-hoc analysis on the best evolved solution for each approach to assess their quality more accurately.
The procedure for post-hoc analysis is similar to the fitness evaluation, with a few key differences.
Training uses 56500 examples instead of 6500 and continues for 1000 epochs instead of 100. 
Moreover, the early stop mechanism is removed.
After training, the weights that achieved the best validation-accuracy are restored for a final test-accuracy assessment.
The system calculates the test-accuracy using 10000 images from the Fashion-MNIST test data.
This test data is a separate data set, never used during evolution, thus ensuring an accurate assessment of the optimizers' ability to generalize to new data.
We repeat the entire procedure 15 times, training the network from scratch and recalculating the test accuracy in each repetition.
After all these steps, the average test-accuracy of an optimizer is the best measure of its actual quality.

\section{Results}
In this section, we report and analyze the results obtained from the 30 runs of each of the four mutation approaches (FMX, FM, OMX, OM).
For a summary of the results, see Table~\ref{tab:resultsummary}, best results for each metric are highlighted in bold.
We conduct a statistical comparison between the different mutational approaches using the t-test (significance level set to $\alpha < 0.05$) and calculate the effect size and p-value \cite{cohen2013statistical}.
A full overview can be found in the supplementary material (Supplementary Material Tables 2 and 3).
In this section's figures, if there are statistically significant differences displayed, we use asterisks to show the significance level and omit the full p-values ($0.05 > p > 0.01$ = *, $0.01 \geq p \geq 0.001$ = **, $0.01 \geq p \geq 0.001$ = ***, $ p < 0.0001$ = ****).
The four different metrics: \textbf{Best fitness}, \textbf{Population fitness}, \textbf{Population diversity} and \textbf{Computational cost} are discussed separately in different subsections. 

\begin{table}
\captionof{table}{Results summary} 
\label{tab:resultsummary} 
\resizebox{\columnwidth}{!}{
\centering
\begin{tabular}[ht]{c|l|l|l|l}
\hline
\multirow{2}{*}{\textbf{Approach}} & 
\multicolumn{4}{c|}{Metrics}\\ 
\cline{2-5} & 
\multicolumn{1}{c|}{\begin{tabular}[c]{@{}c@{}}Best \\ Fitness\end{tabular}} &
  \multicolumn{1}{c|}{\begin{tabular}[c]{@{}c@{}}Population \\ Fitness\end{tabular}} &
  \multicolumn{1}{c|}{\begin{tabular}[c]{@{}c@{}}Population \\ Diversity\end{tabular}} &
  \begin{tabular}[c]{@{}c@{}}Computational \\ Cost\end{tabular} \\  
\hline
FMX & \textbf{0.933[L]} & \textbf{0.625[L]} & \textbf{599[M]} & 2915\\
\hline
FM & 0.931 & \textbf{0.608[L]} & 323 & \textbf{2137[M]}\\
\hline
OM & 0.927 & 0.281 & 122 & 2978\\
\hline
OMX & 0.928 & 0.093 & 73 & 10933\\
\hline
\end{tabular}
}
\small
Values for the four metrics across approaches.
Best results are in bold with Cohen's d  \cite{cohen2013statistical} effect size as "S": small; "M":medium; and "L":large
\end{table}

\subsection{Best Fitness} \label{sec:results_best_fitness}
\begin{figure}
\centering
\includegraphics[width=0.5\textwidth]{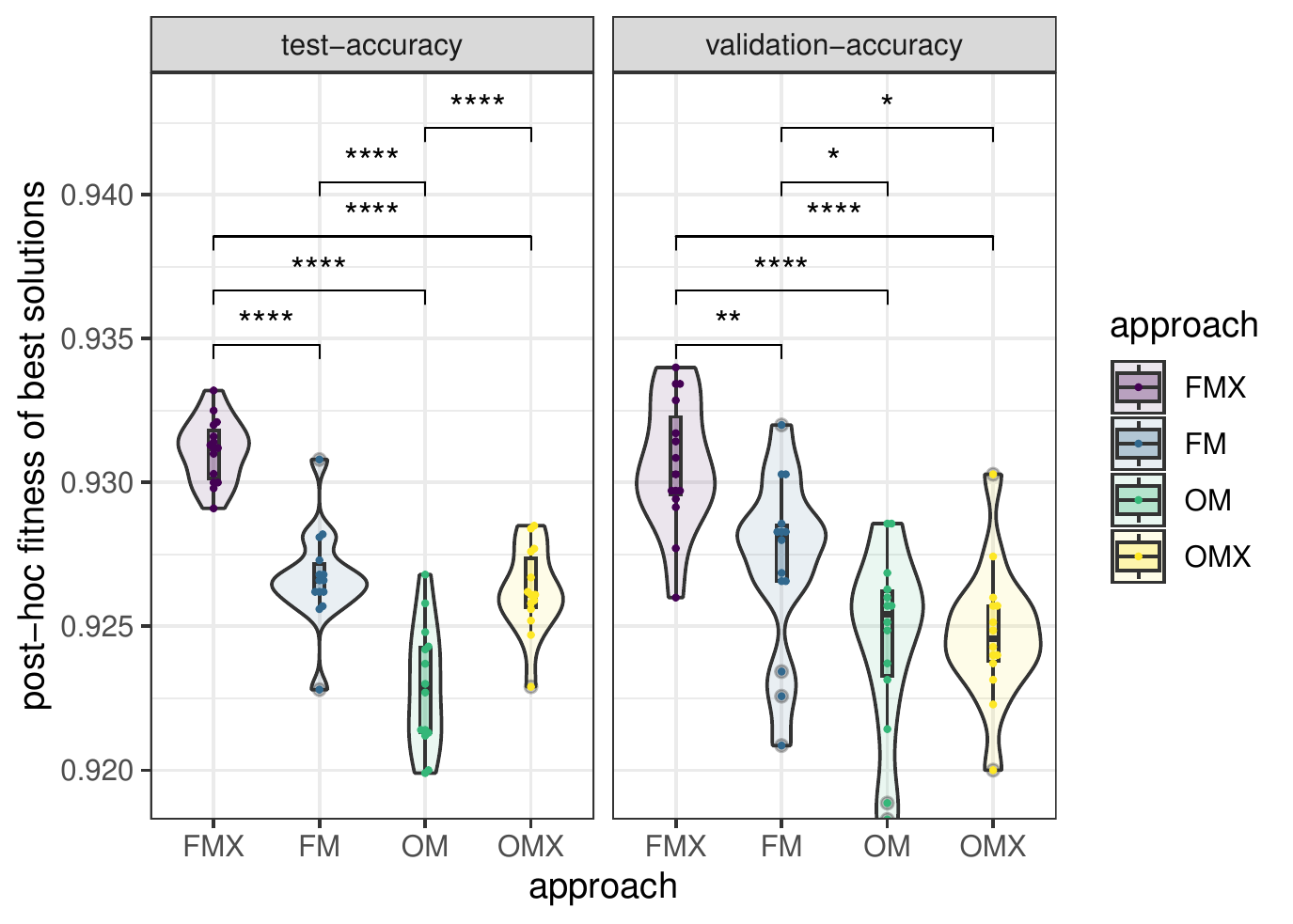}
\caption{Test and validation accuracies of the best optimizer for each setup in post-hoc analysis.}
\label{fig:post_hoc_plot}
\end{figure}

We perform post-hoc analysis on the single best optimizer out of the 30 different runs from each approach and compare their final quality.
The result of the test and validation accuracy assessments, as well as the p-value results from pairwise optimizer comparisons, are shown in Figure~\ref{fig:post_hoc_plot}.
Looking at the results, one can see that the approaches that rely on FM produce the best results overall.

After post-hoc analysis, FMX produced the most effective optimizer, classifying 93.3\% of test images correctly on average.
FMX's optimizer is statistically superior to all other approaches in test and validation-accuracy.
The FM and OMX optimizers are comparable in test-accuracy, despite FM being superior in validation-accuracy.
OM evolved the weakest optimizer.
These results confirm that FM aids evolution in producing superior optimizers.
%FM worked best when combined with a low crossover rate (FMX, crossover rate = 0.01).
%Crossover also plays an important role, as it influences exploration and exploitation, affecting the consistency of evolution in finding high-quality solutions.
Both OM and FM improve when combined with crossover, suggesting that this operator contributes to discovering the best solutions despite its apparent destructive effects.

\begin{figure}
\centering
\includegraphics[width=\columnwidth]{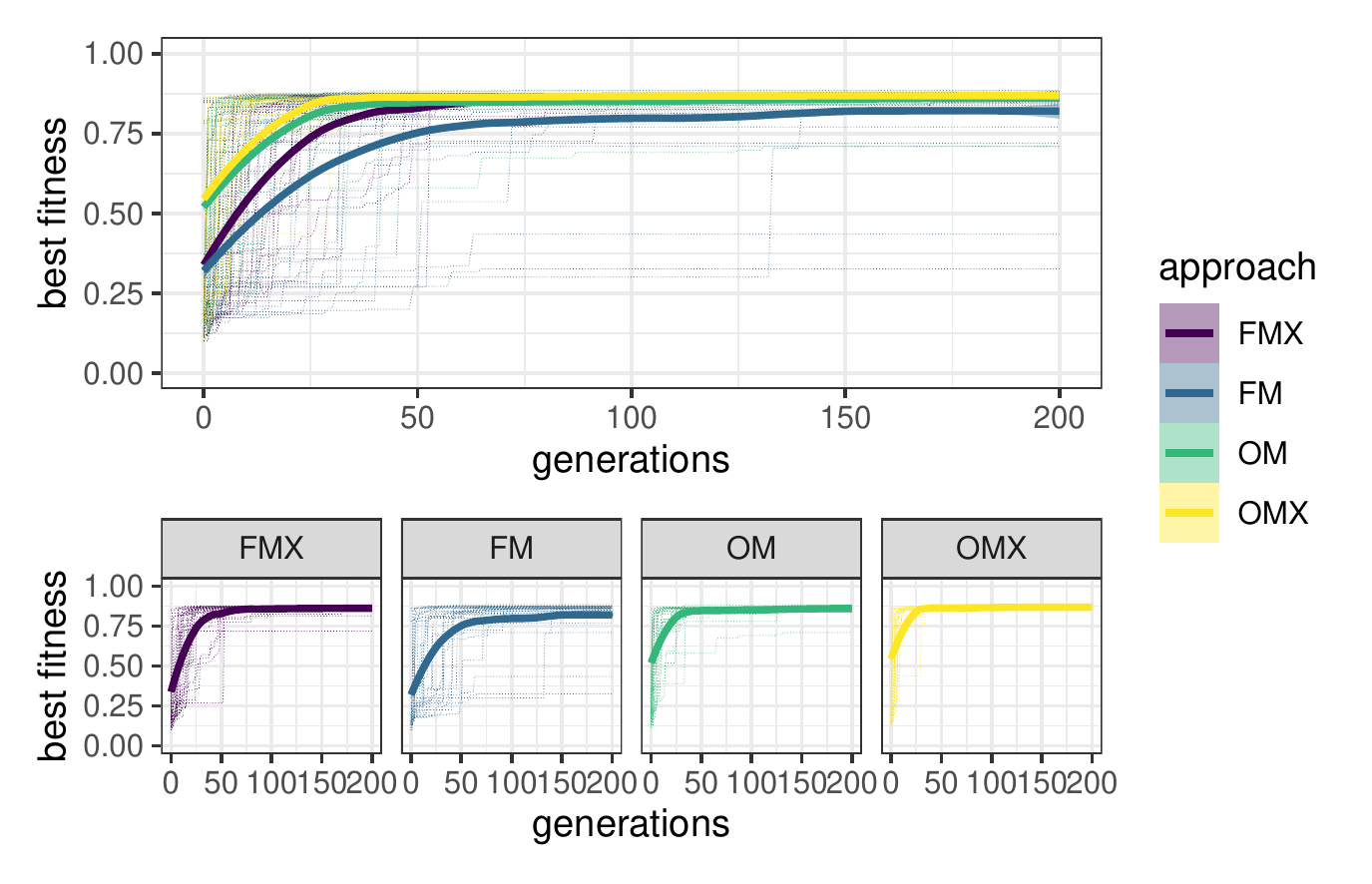}
        \setlength{\belowcaptionskip}{-15pt}  
\caption{Fitness of the best individual of the 30 populations over time for the four different approaches. Bold lines show mean values in each approach.
%Bold lines show the average fitness of the best solution across the 30 runs in each setup over time.
} \label{fig:best_fitness_plot}
\end{figure}

We also analyzed the consistency of the approaches, comparing the best solutions found during evolution by all 30 runs of each approach (shown in Figure~\ref{fig:best_fitness_plot}).
FMX, OM, and OMX have comparable consistency in discovering high-fitness solutions.
OM and OMX methods plateau rapidly as their mutational approaches explore the solution space very fast but impair incremental change, meanwhile FMX reaches a comparable level more gradually by steadily improving upon discovered solutions.
FM is the most inconsistent, with a substantial number of runs below 0.5 fitness for the best solution.

This is due to the absence of crossover combined with the prevalence of neutral mutations, and the use of a solution archive.
Without crossing over, FM mutation can only navigate the solution space via frequent regulatory mutations that do not affect the solution i.e.: neutral mutations.
This evolutionary pattern is called neutral search, and it allows solutions to move across the solution space while being shielded from selection%since they remain selectively the same (neutral) while changing their genotype
. 
Neutral search is computationally efficient in AutoLR because the system only spends time evaluating new solutions and not new genotypes. 
However, since runs for all mutational approaches terminate after a fixed number of generations, FM will explore fewer solutions than the others.
Therefore, FM is penalized by AutoLR
%in terms of exploration
as it has fewer chances of discovering promising optimizers and is more prone to get stuck into sub-optimal regions of the solution space. 
On the other hand, FMX can discover higher-performance solutions thanks to the low rate of crossing over, which nudges the population into unexplored regions of the solution space, where new solutions can be evaluated.
OMX and OM need not rely on the crossover as their highly disruptive mutation approach is sufficient to move them to new regions of the solution space. 

\subsection{Population Fitness}
\begin{figure}[ht]
        \centering
        \includegraphics[width=\columnwidth]{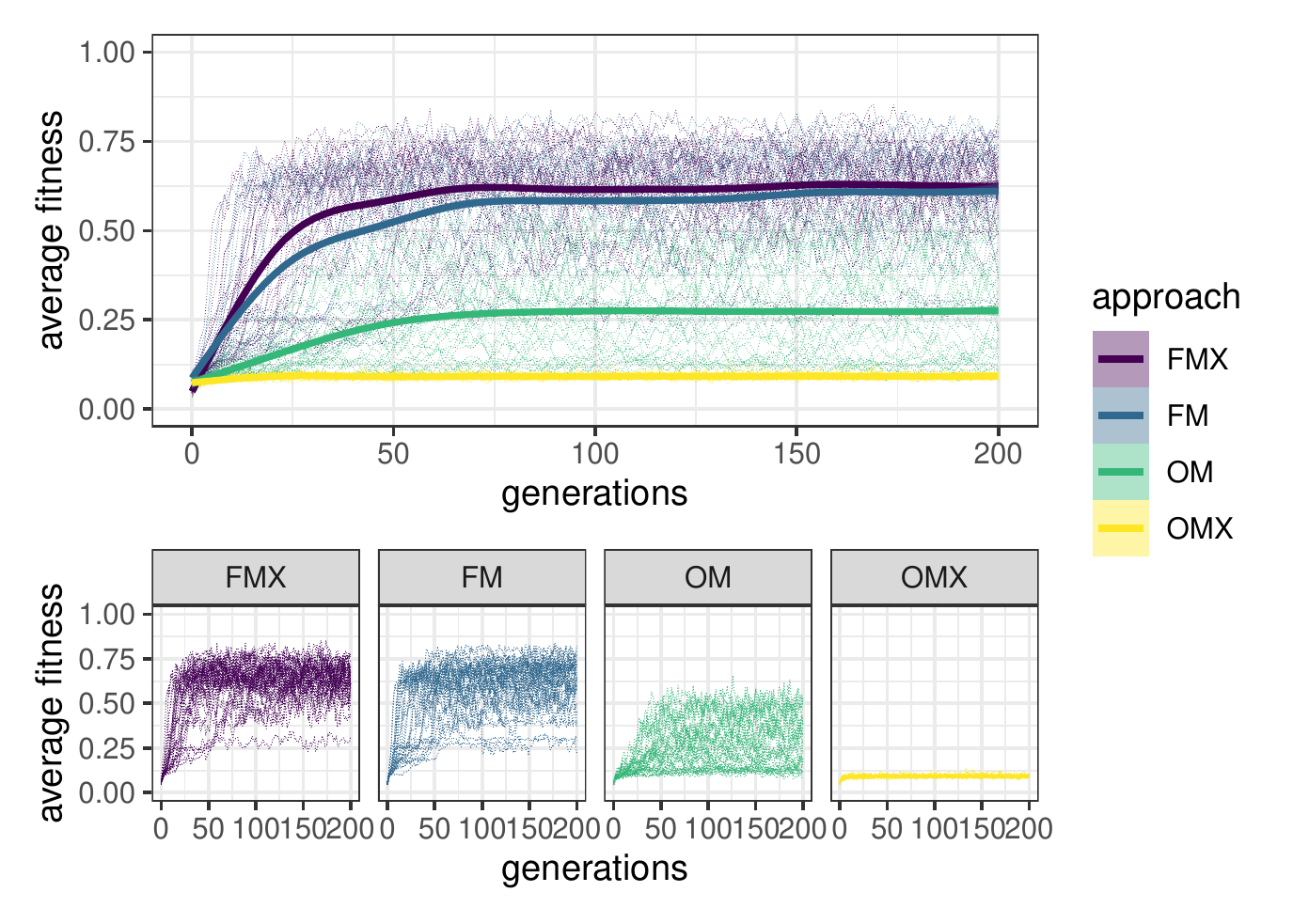}
                \setlength{\belowcaptionskip}{-15pt}  \caption{Average fitness of the 30 populations over time for the four different approaches. }
        \label{fig:average_population_fitness_plot_evolution}
\end{figure}

\begin{figure}
        \centering
        \includegraphics[width=0.5\textwidth]{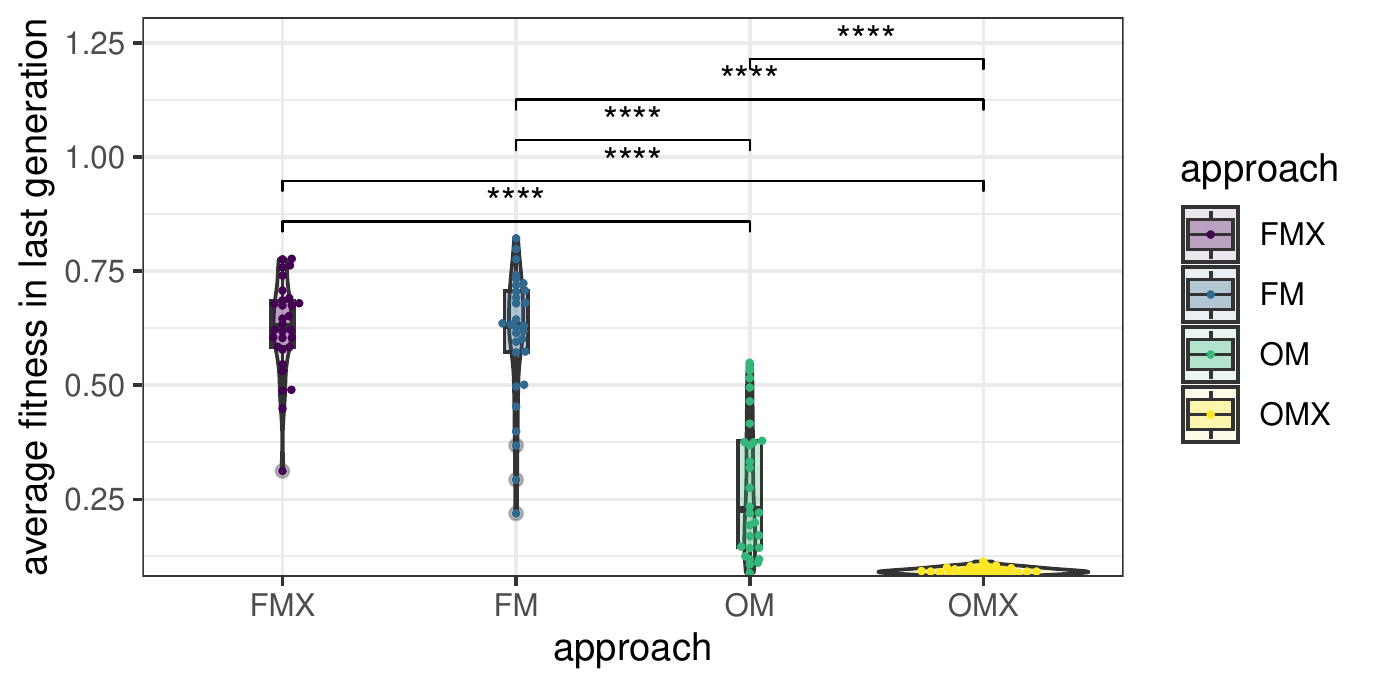}
        \setlength{\belowcaptionskip}{-15pt}  
                    \caption{Average fitness in the last generation of each run for all approaches.}

\label{fig:average_population_fitness_plot_evolution_final}

\end{figure}

FM and FMX substantially improve the average solution fitness in the population. 
At any point in time, FM and FMX populations have higher average fitness than OM/OMX (Figure~\ref{fig:average_population_fitness_plot_evolution}).
FMX and FM have an average population fitness of $\approx$ 0.6, OM and OMX have average fitnesses of $\approx$ 0.3 and $\approx$ 0.1, respectively, as Figure ~\ref{fig:average_population_fitness_plot_evolution_final} shows.
We find that this difference is statistically significant at the end of evolution.

%In previous studies that implemented a regime similar to 
In OMX and OM, average solutions in a population have a fitness one order of magnitude smaller than the best one (OMX and OM average solution fitness $\approx$ 0.1, OMX and OM best solution fitness $\approx$ 0.9).
This is in accordance with results from previous studies\cite{Carvalho2022}.
In contrast to FMX, OMX proceeds by randomly sampling good solutions from a vast solution space thanks to its uniform mutation rate and high crossover rate (0.9) but does not traverse the search space efficiently, inhibiting fit solutions from producing incrementally fitter offspring.
%We compare the average population fitness throughout evolution (shown in %Figure~\ref{fig:average_population_fitness_plot_evolution}).
%Additionally, we statistically compare the average population fitness at the %end of each run in Figure~\ref{fig:average_population_fitness_plot_evolution_final}).
These results demonstrate the delicate role of crossover rate in balancing exploration and exploitation and the importance of tuning this parameter correctly.
A high crossover rate has a highly negative impact on population fitness. Population fitness in OMX (crossover rate  = 0.9) is significantly lower than its crossover-free counterpart OM (crossover rate  = 0).
This difference contrasts with the positive effect that small crossover rates can have, as observed in Section~\ref{sec:results_best_fitness}. 
FMX and FM population fitnesses do not show the same drop.
%This difference in response to crossover is caused by the difference in crossover rate between FMX and OMX (FMX crossover rate $= 0.01 <<$ OMX crossover rate $=$ 0.9).
The difference in the response to crossover may caused by the different crossover rates (FMX crossover rate $= 0.01 <<$ OMX crossover rate $=$ 0.9).
Further study into how each approach reacts to changes in crossover rate is necessary to confirm this idea since such experiments are outside the scope of this work.

Improved population fitness supports the hypothesis that FM and FMX approaches can improve the navigation of the solution space.
However, analyzing the population's fitness alone is not sufficient.
Many identical copies with high fitness could lead to the same result.
It is essential to analyze the diversity of solutions discovered by each approach to asses if FM and FMX approaches can explore the solution space better.

\subsection{Population Diversity} \label{sec:results_diversity}
We measure population diversity as the \textit{number of viable unique behaviors in the population}. 
The metric we use is a modified version of the \textit{variety} measure proposed by Koza \cite{koza1992a}. This measure is defined as "the percentage of individuals for which no exact duplicate exists elsewhere in the population" we adapt this to work on \textit{number of unique behaviors} rather than percentages but retain the focus on "individuals for which no exact duplicate exists elsewhere in the population" which we abbreviate to "unique behaviors in the population". We could use an edit-distance based measure instead, but we find that a simple, behavior focused measure is the best match for this study. 

Optimizers are fragile and non-viable solutions are frequent, thus including non-viable solutions in our analysis would confound information about the diversity of optimizers that actually contribute to evolution. 
We find "the number of \textit{viable} unique behaviors" by considering solutions with a fitness above the threshold of 0.5, that are able to train the network in some capacity. 
We statistically compare the total number of unique viable behaviors discovered per run (Figure~\ref{fig:pop_diversity_plot}, for the number of unique behaviors regardless of fitness see Supplementary Material Figure 1).
\begin{figure}
        \centering
        \includegraphics[width=\columnwidth]{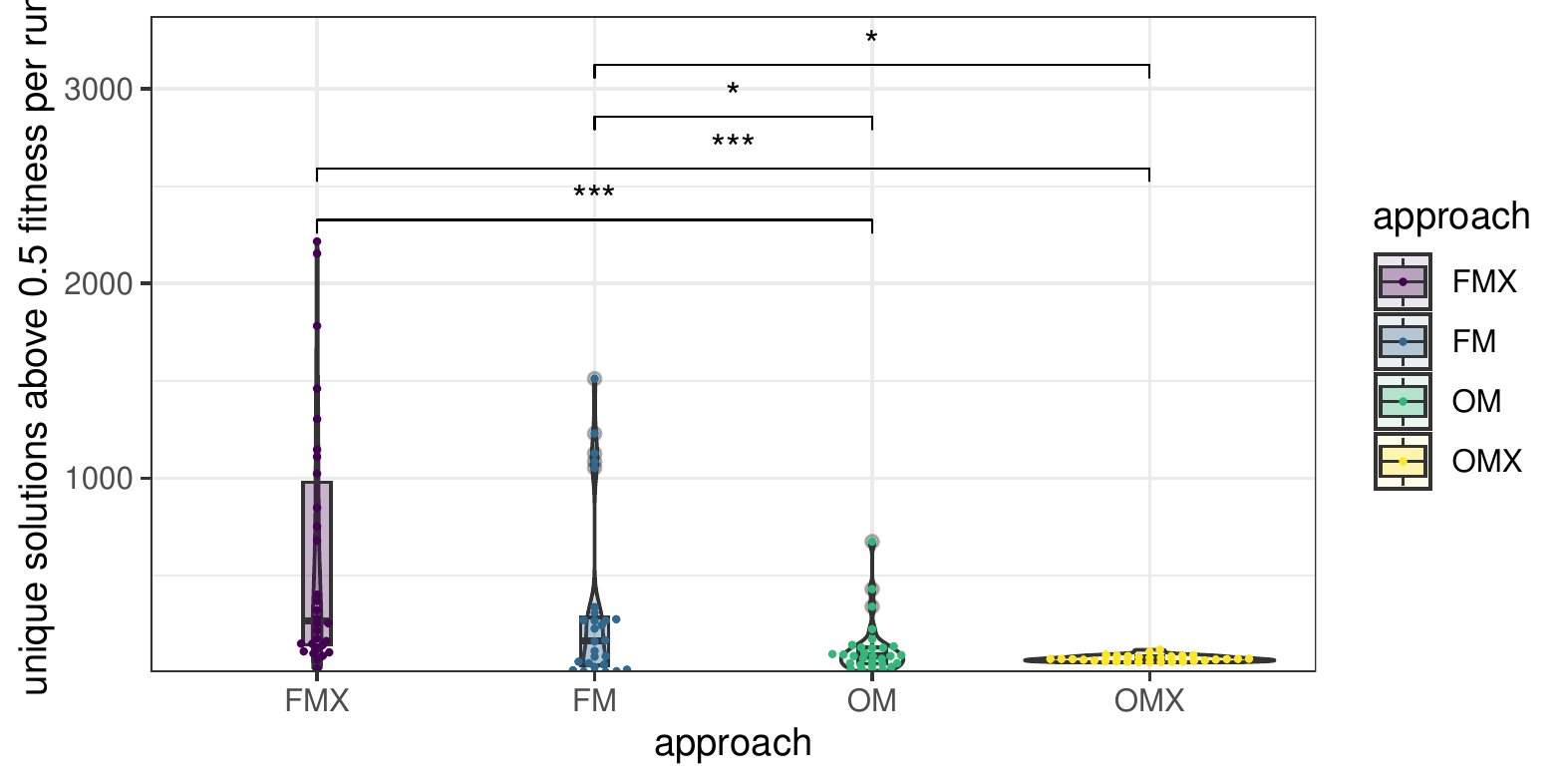}
        \caption{Unique behaviors above 0.5 fitness discovered per run. Higher values mean more diversity.\\}
        \label{fig:pop_diversity_plot}
\end{figure}

FMX produces a significantly higher diversity of viable solutions than all other approaches (on average 600 viable unique solutions per run). FM produces on average 300 and OM 120, although this difference is not statistically significant, while OMX is statistically worst than all other approaches with just 70 unique viable solutions per run on average (Figure~\ref{fig:pop_diversity_plot}). 

The lack of diversity in OMX clues us into the failures of this approach and confirms our hypothesis that FMs and FMX can explore the solution space more efficiently.
Despite the vast amount of new behaviors that crossover introduces into the population, OMX cannot convert this into functioning behaviors that aid evolution.
The combined insights about diversity and population fitness suggest that FMX and FM can keep a diverse, viable population longer than OM and OMX.
In sum, these experiments suggest that FM and FMX succeed in the stated goal of improving the search for the solution space, possibly allowing them to avoid 
%local minima and 
stagnation.

However, the results raise questions about performance.
We previously hypothesized that FM might be computationally cheaper by evaluating fewer solutions thanks to an increase in neutral search.
This analysis of diversity suggests that FM evaluates more viable solutions.
One possible explanation for this result is that FM is saving resources on evaluations that are not reflected in our measure of diversity.
Specifically, FM may reduce the number of evaluations below the threshold used to analyze diversity.
To clarify this aspect, we analyze the computational cost of the experiments directly.

\subsection{Computational Costs}
\begin{figure}[h]
\centering
  \makebox[\columnwidth]{\includegraphics[width=\columnwidth]{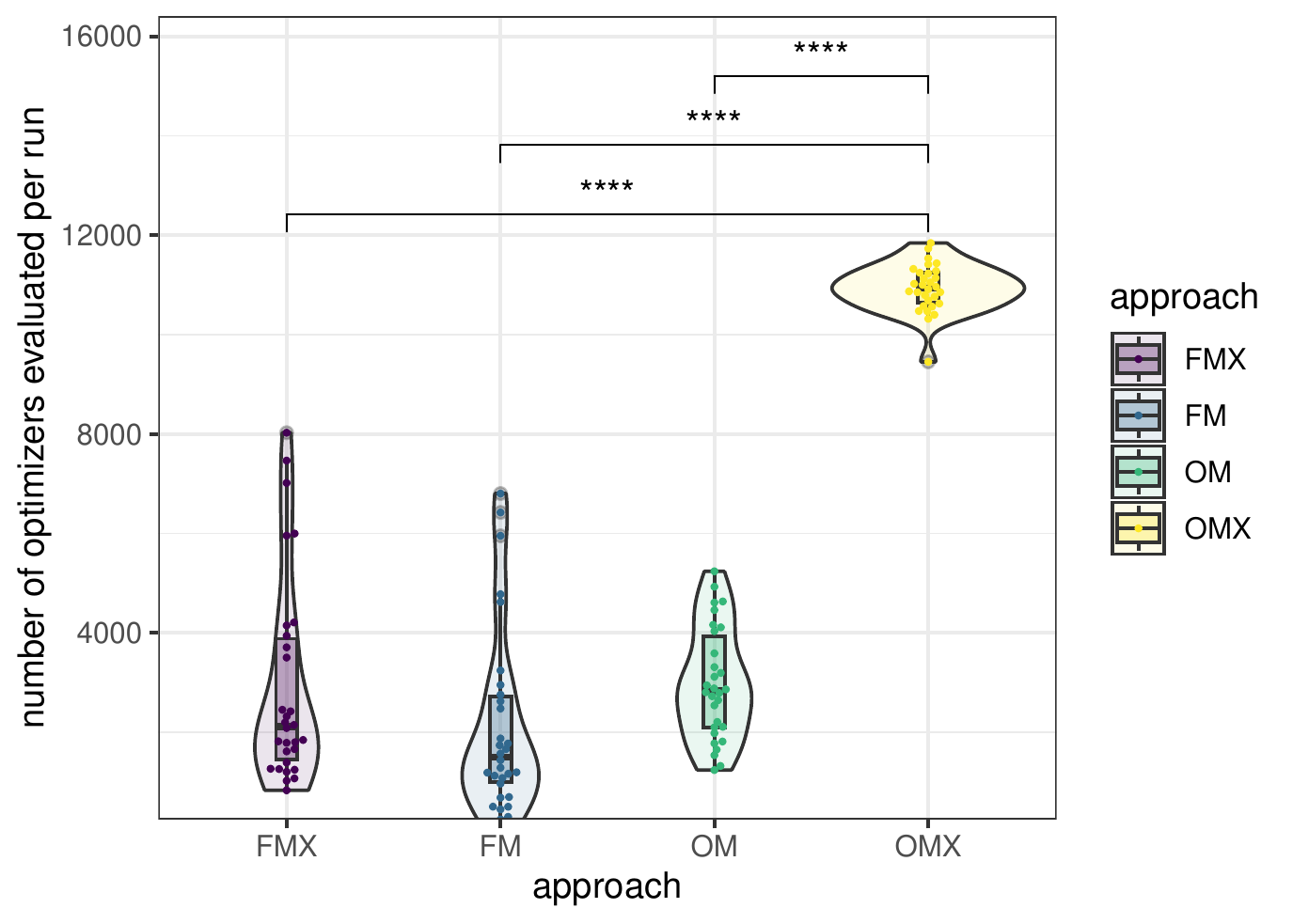}}
  %\makebox[\columnwidth]{\includegraphics[width=\columnwidth]{figures/number_of_cumulative_unique_solutions_with_fitness_above_0.5_over_time_per_run_across_regimes.pdf}}
\setlength{\belowcaptionskip}{-20pt}  
\caption{Total number of evaluated optimizers, for all approaches. Lower values mean less computational costs.}
\label{fig:computational_cost_plot}
\end{figure}

Most of the computational costs of AutoLR come from the neural network training necessary for the fitness evaluation.
In Figure~\ref{fig:computational_cost_plot}, we statistically compare the number of training and evaluations (the main computational cost) executed in all approaches.
AutoLR uses a solution archive and pre-selects only for functional optimizers to avoid redundant and irrelevant evaluations. 
Due to these mechanisms, the number of optimizers evaluated in a run is the number of unique behaviors that include at least one gradient component.
All approaches we have implemented for the first time in this work (FMX, FM, OM) are statistically cheaper in terms of computational cost than previously implemented ones (OMX), see Figure~\ref{fig:computational_cost_plot}).
There is no statistically significant difference between FMX, FM, and OM (although FM sits just outside the established significance threshold when compared to OM, with a p-value = 0.057, see Supplementary Material Table 3).
Moreover, the results seem to indicate that FM is empirically cheaper than all the other approaches, but this difference is not statistically significant.
FM thus seems to manage to be cheaper while at the same time having a high diversity of good optimizers (see Section~\ref{sec:results_diversity}).
This suggests that FM performs neutral search (saving computational resources) in high-fitness regions of the problem space.

\section{Conclusions}

This paper proposes Facilitated Mutation, a mutational approach for GGGP inspired by biological evolution.
Facilitated mutation splits the single mutation rate 
commonly used in these approaches
into different values for each grammar non-terminal.
We compare Facilitated Mutation approaches with (FMX) and without crossover (FM)  against more traditional mutational approaches with (OMX) and without (OM) crossover. We validate these approaches on optimizer evolution, an evolutionary machine learning task with %fragile solutions,
a challenging solution space, and real-world applications%.
, and compare them
%We compare the approaches 
based on the best optimizer discovered, average population fitness, population diversity 
%(measured as the number of unique viable solutions discovered)
, and computational cost %(measured as the number of fitness evaluations)
.

%We find that 
FMX is comparable or better to original mutation approaches in all aspects.
When compared to the best OM-based alternative: FMX evolved a better optimizer (+0.5\% network test accuracy than OMX 
%,+0.06\% than OM
), and on average: produced better and discovered more solutions (+34\% population fitness, +477 viable solutions)
than OM
%, +0.34\% than OM
%),  viable solutions ( on average than OM
%, +477 than OM
%).
These benefits come with no significant downsides, the approach has similar computational costs to OM and is significantly cheaper than OMX (saving $\approx$ 118000 fitness evaluations% on average
).

Comparing the two mutation approaches without crossover (FM vs. OM), we find that FM discovered a better optimizer (+0.4\% network test accuracy), improved population fitness by +32\%, discovered 200 additional viable individuals %on average
and decreased computational costs by 850 fitness evaluations on average.

While we only study its benefits in the context of an SGE-based framework, FM is %theoretically 
applicable to any GGGP where each gene is directly associated with a grammar non-terminal. %, and any GP solvable problem.
%Furthermore, the approach is not tied to the problem we address in this work.
FM is simple in implementation and applicable to any GP problem, but the effectiveness depends on the type of problem and the choice of proper grammar and mutation rates.

\subsection{Future Work}
FM and FMX achieve several remarkable results in this work, but these mutational approaches can still be pushed further.

In this work, the rates of mutation and crossover have not been further fine-tuned after their initial choice.
There are many ways to further optimize mutation rates for FM; most notably, we think a promising direction is to let the mutation rates themselves evolve% to avoid the need for manual tuning
. Since the submission of this work, we have created a new version of FM with adaptive mutation rates and outlined a grammar design procedure to amplify the benefits of this type of mutation \cite{carvalho2023context}.

Our analysis of crossover in this work can also be expanded.
We focused on assessing the benefits of our solution through an informed decision of crossover rate (FMX crossover rate = 0.01) and drawing comparisons with another work with a comparable setup (OMX crossover rate = 0.9).
However, we do not investigate how the different mutation approaches respond to different crossover rates.
Crossover significantly improved the best fitness of both approaches so it is essential to study how these methods respond to varying crossover rates.

The analysis of the different mutational approaches can also be expanded to understand all their implications better.
While we found viable measures for all aspects of the evolution we studied, technical limitations stopped us from looking at other metrics, such as solution complexity, the phenotypic distance between solutions, and the total run time of experiments.

Other aspects of evolution could also be studied, such as detecting optimizers that evolution discovers repeatedly and how these vary depending on the approach.
The improved diversity and population fitness of FM approaches also warrant further research into whether the approach can also improve the system's evolvability and adaptation to different networks and tasks through transfer learning. In particular we find important to evaluate the performance of our approach in more challenging image recognition data sets (e.g., CIFAR-10). From our first analyses it appears that the performance for the Fashion-MNIST data set is saturated (see \ref{fig:best_fitness_plot}). FM methods may show an even larger margin of improvement in more complex data sets, where simpler mutational regimes (OM) could not navigate the higher complexity of the search space.  

FM may improve performance even further for structured problems (e.g., robot control, path finding) where there is a limited set of possible actions that can be fully defined by the grammar. This contrasts with optimizer evolution, where the set of variables and operations used for evolution is arbitrarily restricted, infinitely varied and endlessly debatable.

Finally, previous works compared the evolved optimizers with human-made solutions.
We did not perform this type of analysis since the focus of this work was on improving evolution, but such insights would be interesting in an extended version of this paper.

\section{Acknowledgements}
We thank the members of the modelling adaptive response mechanisms (MARM) group at the University of Groningen for stimulating discussion and comments. S.T., acknowledge funding from the European Research Council (ERC Advanced Grant No. 789240).
This work was also funded by FEDER funds through the Operational Programme Competitiveness Factors - COMPETE and national funds by FCT - Foundation for Science and Technology (POCI-01-0145-FEDER-029297, CISUC - UID/CEC/00326/2020) and within the scope of the project A4A: Audiology for All (CENTRO-01-0247-FEDER-047083) financed by the Operational Program for Competitiveness and Internationalisation of PORTUGAL 2020 through the European Regional Development Fund. The second author is funded by FCT, Portugal, under the grant UI/BD/151053/2021.

\bibliographystyle{ACM-Reference-Format}
\bibliography{references}

\end{document}